\documentclass[lettersize,journal]{IEEEtran}
\usepackage[numbers]{natbib}
\usepackage{caption}
\usepackage{setspace}
\usepackage{amssymb}
\usepackage{multirow}
\usepackage{lscape}
\usepackage{amsthm}
\usepackage{booktabs}       
\usepackage{amsmath}        
\usepackage{mathrsfs}
\usepackage{longtable}
\usepackage{url}
\usepackage{tabularx}
\usepackage{epsfig}
\usepackage{subcaption}
\usepackage{graphicx}
\usepackage{textcomp}
\DeclareCaptionLabelFormat{cont}{#1~#2\alph{ContinuedFloat}}
\usepackage{threeparttable}
\usepackage{array}
\usepackage{lipsum}         
\usepackage{wrapfig}
\usepackage{epstopdf}
\usepackage{bm}
\usepackage{makecell}
\usepackage{stfloats}
\usepackage{diagbox}
\usepackage{float}
\usepackage{color}
\usepackage{algorithm}
\usepackage{algorithmicx}
\usepackage{algpseudocode}
\usepackage{tcolorbox}
\usepackage{microtype}      
\usepackage{hyperref}
\usepackage{braket}
\usepackage{amscd}
\usepackage{algorithm}
\usepackage{algorithmicx}
\usepackage{algpseudocode}

\newtheorem{definition}{Definition}

\begin{document}

\title{Visual Evolutionary Optimization on Graph-Structured Combinatorial Problems with MLLMs: A Case Study of Influence Maximization}

\author{Jie Zhao, Kang Hao Cheong,~\IEEEmembership{Senior Member, IEEE}
\thanks{
This work was supported by the Singapore Ministry of Education (MOE) Science of Learning, under Grant No. MOESOL2022-0003.

Jie Zhao and Kang Hao Cheong are affiliated with the Division of Mathematical Sciences, School of Physical and Mathematical Sciences, Nanyang Technological University, S637371, Singapore. Kang Hao Cheong is also with the College of Computing and Data Science, Nanyang Technological University, S639798, Singapore.

Corresponding Author: K.H. Cheong (kanghao.cheong@ntu.edu.sg).}
}

\markboth{Journal of \LaTeX\ Class Files,~Vol.~14, No.~8, August~2021}%
{Shell \MakeLowercase{\textit{\textit{et al.}}}: A Sample Article Using IEEEtran.cls for IEEE Journals}


\maketitle

\begin{abstract}

Graph-structured combinatorial problems in complex networks are prevalent in many domains, and are computationally demanding due to their complexity and non-linear nature. Traditional evolutionary algorithms (EAs), while robust, often face obstacles due to content-shallow encoding limitations and lack of structural awareness,  necessitating hand-crafted modifications for effective application. In this work, we introduce an original framework, visual evolutionary optimization (VEO), leveraging multimodal large language models (MLLMs) as the backbone evolutionary optimizer in this context. Specifically, we propose a context-aware encoding scheme, representing the solution of the network as an image. In this manner, we can utilize MLLMs' image processing capabilities to intuitively comprehend network configurations, thus enabling machines to solve these problems in a human-like way. We develop MLLM-based operators tailored for various evolutionary optimization stages, including initialization, crossover, and mutation. {Furthermore, we propose that graph sparsification can effectively enhance the applicability and scalability of VEO on large-scale networks, owing to the scale-free nature of real-world networks.} We demonstrate the effectiveness of our method using a well-known task in complex networks, influence maximization, and validate it on eight different real-world networks of various structures. The results confirm VEO's reliability and enhanced effectiveness compared to traditional evolutionary optimization.

\end{abstract}

\begin{IEEEkeywords}
Multimodal large language models, evolutionary optimization, influence maximization, complex networks
\end{IEEEkeywords}

\section{Introduction}
\IEEEPARstart{G}{raph}-structured combinatorial problems are pervasive across various fields, such as influence maximization \cite{ma2022influence} in social media. These problems are characterized by their reliance on networks where nodes and edges represent discrete entities and their interactions, respectively. The complexity of these networks, coupled with their typically large and complex nature, poses significant computational challenges \cite{ma2021enhancing,zhao2023obfuscating}. Evolutionary algorithms (EAs) are well-suited for graph-structured problems as they are adept at managing the nonlinearities and discontinuities \cite{zhang2022search,biswas2022improved,pan2021bi}. While traditional EAs have been instrumental in tackling a variety of optimization problems \cite{li2024evolutionary,li2023deep}, they still exhibit an inherent limitation, particularly when applied to graph-structured combinatorial problems due to the encoding limitations. 

The choice of encoding scheme plays a crucial role in solving discrete problems, as it determines how solutions are represented and manipulated within the evolutionary optimization framework. Different discrete tasks require tailored encoding strategies in EAs, such as permutation and path encodings for scheduling \cite{hart2005evolutionary} and routing problems \cite{stodola2022adaptive}, or binary and label-based (node or edge index) encodings for subset selection problems, such as influence maximization \cite{wang2021identifying} and community deception \cite{zhao2025multi}. However, these traditional encodings represent nodes or paths in an abstract manner, failing to capture their roles within the network. As a result, evolutionary operators like crossover and mutation are applied without an inherent understanding of the underlying entities or relationships. This lack of structural awareness can lead to naive modifications that may not respect the structural properties of the graph.

Multimodal large language models (MLLMs) are extensively trained on diverse datasets, enabling them to effectively process multiple data types with inherent adaptability \cite{yin2024survey,hao2022iron}. This capability presents a unique opportunity to enhance evolutionary optimization fundamentally. In this work, we introduce a novel encoding methodology that visually represents solutions as images. By leveraging this visual encoding strategy, we unlock the potential of MLLMs in evolutionary optimization and propose an original framework called \textbf{Visual Evolutionary Optimization (VEO)}. This framework enables the model to analyze and manipulate solutions in a way that mirrors human decision-making in evolutionary operations, such as crossover and mutation. {The motivations and advantages behind VEO operate on two levels as follows:}

\indent \textit{{(1) Representation-level: Image-based Encoding Scheme}}
\begin{itemize}
    \item {\textbf{Expressing High-Order and Global Structural Information:} Image representations intuitively capture complex structural information in graphs, such as indirect relationships and community structures. This visual encoding provides a clear and compact view of high-level features that are difficult to express with traditional representations.}
    \item {\textbf{Seamless Compatibility with MLLMs:} By transforming combinatorial graph problems into visual formats, we enable a seamless pipeline for applying MLLMs to domains traditionally dominated by symbolic or algorithmic methods and unlock the reasoning capabilities of MLLMs for combinatorial tasks.}
    \item \textbf{Scalability and Inference Efficiency:} Image-based representations maintain a consistent input size regardless of graph complexity, thereby avoiding issues such as token explosion, which commonly arises when representing large-scale networks in textual form. This characteristic enables more efficient inference and enhances scalability within MLLM frameworks.
\end{itemize}

\textit{{(2) Optimization-level: MLLMs as Evolutionary Optimizers}}

\begin{itemize}
    \item {\textbf{Human-Like Visual Reasoning:} MLLMs can interpret the structure of image-encoded graph and reason about solution quality in a human-like manner, enabling high-quality decision-making for offspring generation within the EA framework without relying on random search or complex heuristics.}
    \item {\textbf{Generalizability and Adaptability:} MLLMs can adapt across diverse and dynamic graph structures without requiring problem-specific algorithm design. They can also flexibly respond to changes in topology, offering broad applicability and real-time adaptability.}
    \item {\textbf{Constraint-Aware Optimization:} MLLMs can naturally avoid infeasible solutions by understanding constraints through contextual cues, reducing the need for manual constraint handling and dedicated adjustment.}
\end{itemize}


Effective visualization is essential for utilizing VEO on graph-structured problems, particularly when dealing with large networks. Plotting all nodes on a limited canvas often results in cluttered visuals, obscuring critical structural details and connections. Node labels, essential for MLLM-based operations like initialization, crossover, and mutation, must remain clear and readable to ensure that MLLMs can accurately identify each node's role and relationship within the network. {As noted in \cite{barabasi2009scale}, many real-world networks exhibit a scale-free structure, where a small subset of nodes holds disproportionately high centrality or importance. By concentrating on these key nodes and reducing overall graph complexity without significantly compromising solution quality, we are able to address large-scale problem instances more effectively. To this end, we propose a community-level graph sparsification method that compresses large networks into more manageable scales while preserving essential structural information necessary for meaningful analysis and optimization.}

A common challenge with existing community detection algorithms is their tendency to produce an excessive number of communities, often including single-node communities \cite{clauset2004finding}. While this level of granularity could be informative, it is often impractical for graph sparsification. Recognizing the importance of community information \cite{kojaku2024network,chen2019ga} and the need to balance detail with usability, we develop a new community merging method to reduce the number of communities while preserving the essential structural information during graph sparsification. The main contributions of our work are summarized as follows:

$\bullet$ Our paper presents a novel context-aware encoding scheme for graph-structured combinatorial optimization, where solutions are represented as images that encapsulate rich structural information of the underlying networks. The proposed representation allows us to further develop evolutionary operations with a profound understanding of the network's topology. More precisely, we introduce MLLM-based operators tailored to various stages of evolutionary optimization, including initialization, crossover, and mutation, fundamentally redefining how solutions are generated, manipulated, and refined within the evolutionary framework.

$\bullet$ We explore how to effectively improve the applicability and scalability of MLLMs by leveraging graph reduction techniques. Specifically, we propose a community-level sparsification method that reduces the size of large-scale networks while preserving key structural information. As part of this process, a new community merging technique is employed to consolidate close communities, enhancing sparsification efficiency while maintaining meaningful graph semantics.

$\bullet$ We have comprehensively investigated the reliability of VEO, such as the validity and effectiveness of MLLMs output. Several key parameters such as the influence of the number of nodes in the plotted network as well as the different-level tasks on the quality of MLLMs are also studied. The experiments are conducted on eight different real-world networks and the ablation results suggest the proposed VEO can greatly facilitate different stages of evolutionary optimization.

{As an illustration, we adopt the influence maximization problem as a representative case study to illustrate the potential of MLLMs as effective evolutionary operators. To maintain generalizability and focus, our approach adheres to the standard evolutionary optimization framework in this work.}

The structure of this paper is as follows: Section \ref{sec.related_work} discusses previous research on evolutionary optimization, LLMs and combinatorial problems. Section \ref{sec.MLLM} details the proposed MLLM-driven evolutionary optimization. In Section \ref{sec.experiment}, we evaluate the effectiveness of VEO and analyze various parameters. The future work for the MLLM-based evolutionary optimization on graph-structured combinatorial problems is discussed in Section \ref{sec.future}. Finally, Section \ref{sec.conclusion} presents the conclusions of the paper.

\section{Related work}\label{sec.related_work}
In this section, we will review some literature related to evolutionary optimization, combinatorial problems and large language models.

\subsection{Evolutionary Optimization on Combinatorial Problems}
The utility of evolutionary optimization in addressing discrete and non-linear problems has prompted its widespread adoption for tackling complex networks \cite{wang2019surrogate,qiu2024scalable} and various combinatorial tasks \cite{wang2023multiobjective,liu2023evolutionary}, such as trunk scheduling \cite{chen2024deep,chen2022cooperative} and job shop scheduling \cite{yao2024bilevel,zhang2022multitask}. In the realm of complex networks, the versatility of evolutionary optimization is further demonstrated by its widespread use in enhancing network robustness \cite{wang2021computationally,wang2023enhancing} and facilitating network reconstruction \cite{wu2020evolutionary,ying2021multiobjective}. Additional applications include identifying critical nodes \cite{zhang2023interactive}, discovering dynamic communities \cite{ma2023higher}, recognizing network modules \cite{chen2019mumi,gao2023multilayer} and sensor deployment \cite{zhao2025enhanced}, thereby illustrating its broad applications in network-related challenges. 


\subsection{LLMs-Assisted Evolutionary Optimization}
Wu \textit{et al.} \cite{wu2024evolutionary} classified existing works regarding the synergy of LLMs and evolutionary computation into two main branches: LLMs-based search operators \cite{baumann2024evolutionary,brahmachary2024large} and LLMs-based algorithm automation \cite{morris2024llm,van2024llamea,sobania2024comparison}. Our study can  be roughly classified into the first category. In 
\cite{meyerson2023language}, LLMs are employed as crossover operators to derive new solutions from parental inputs. Brownlee \textit{et al.} \cite{brownlee2023enhancing} also presented LLMs effectively functioning as mutation operators that enhance the search process. Liu \textit{et al.} \cite{liu2023large} introduced a novel framework known as LLMs-driven EA (LMEA), which utilizes LLMs for both crossover and mutation operations. Furthermore, LLMs-based search operators were adapted for multi-objective scenarios by segmenting traditional optimization tasks into sub-problems \cite{liu2023large1}. The related application can be found in \cite{tran2024proteins}, where protein properties are optimized with LLMs-enhanced evolutionary operators.

\subsection{LLMs on Graph-Structured Problems}

The recent advancements in LLMs have led to significant progress across diverse fields including sentiment analysis \citep{deng2023llms}, optimization \citep{romera2024mathematical}, and social sciences \citep{zhang2024toward}. This raises the question of whether the capabilities of LLMs can be extended to graph-related tasks \citep{chen2024exploring,tang2024higpt}. However, as indicated by studies such as \citep{fatemi2023talk,wang2024can}, LLMs struggle with graph-structured data and often fail to achieve satisfactory results on fundamental tasks, especially as graph complexity increases. 


{Recent studies have explored the application of MLLMs to graph-related tasks, primarily as benchmarks to evaluate their visual reasoning capabilities. Wei \textit{et al.} \cite{wei2024gita} combined visual and textual representations of graphs to enable MLLMs to perform structured reasoning, while VisionGraph \cite{li24ab} relied exclusively on visual inputs to assess how well these models interpret and reason over graphical structures. Beyond basic graph reasoning tasks, there are also some applications of MLLMs for combinatorial optimization. Huang \textit{et al.} \citep{huang2024multimodal} proposed a multimodal approach that combines visual and textual inputs to solve the traveling salesman problem (TSP), whereas Elhenawy \textit{et al.} \citep{elhenawy2024eyeballing} focused solely on visual representations to address routing tasks using MLLMs. Although both studies represent pioneering efforts in applying MLLMs to combinatorial problems, they do not incorporate evolutionary optimization frameworks and are limited to relatively small-scale networks with fewer than 200 nodes.}

Our work is, to the best of our knowledge, the first to integrate MLLMs as evolutionary optimizers for solving large-scale combinatorial problems, with a particular focus on scalability and structural understanding of real-world complex networks. We aim to explore this untapped synergy by investigating how MLLMs can facilitate structure-aware evolutionary optimization in complex network scenarios, using influence maximization as a representative case study.

\section{MLLM-driven evolutionary optimization}\label{sec.MLLM}
In this section, we first introduce a classical combinatorial problem to optimize \cite{wang2023multia,wen2024eriue}, followed by detailed presentations of the proposed graph sparsification and community merging methods, along with the MLLM-based evolutionary optimizer.

\subsection{Problem Formulation}
In this work, we illustrate the proposed VEO with influence maximization \cite{hong2020efficient,liu2024community}. Consider a network \( G = (V, E) \), where we aim to identify a subset of nodes \( S \subseteq V \) that maximizes the spread of influence throughout the network and the objective is expressed as:
\begin{equation}
\max_{S \subseteq V, |S| = k} \sigma(S),
\end{equation}
where \( \sigma(S) \) is the expected influence spread of the seed set \( S \), and \( k \) is the predefined number of nodes of \( S \).

Due to the computational intensity of running Monte Carlo methods for diffusion simulation, we use the widely adopted Expected Diffusion Value (EDV) \cite{jiang2011simulated} as the fitness function to evaluate diffusion speed. For a small propagation probability \( p \), the EDV estimates the expected number of nodes influenced by a seed set \( S \) as:
\begin{equation}
EDV(S) = k + \sum_{b \in \mathcal{N}(S)/S} \left(1 - (1-p)^{\delta(b)}\right),    
\end{equation}
where \( \mathcal{N}(S) \) represents the one-hop neighbors of the seed set \( S \), defined as \( \mathcal{N}(S) = S \cup \{b \mid \exists s \in S, (s, b) \in E\} \), and \( \delta(b) \) denotes the number of edges connecting \( b \) to any node in \( S \).

\subsection{Graph Sparsification}
The applicability of MLLMs would be challenged when attempting to visualize large-scale networks on a single canvas, where displaying all nodes with labels can lead to significant overlaps and a chaotic structure, making direct analysis and optimization impractical. As indicated in \cite{barabasi2009scale}, many real-world networks are scale-free, meaning that only a minority of nodes carry high centrality or importance. By focusing on these key nodes and reducing graph complexity without significant loss of solution quality, we are able to explore an approach applicable to large and real-world problem instances. To this end, we resort to the sparsification technique that simplifies the original network while preserving its essential topological features.  Let the original network be represented by $G = (V, E)$, its simplified network \( G_s \) can be obtained using a sparsification operator \( \mathcal{S}(\cdot;\theta) \), specifically,
\begin{equation}
    G_s = \mathcal{S}(G; \theta),
\end{equation}
where \( \theta \) refers to the specific graph sparsification technique and it satisfies $G_s \subset G$.

Consider a graph $G = (V, E)$ with a set of communities $\widetilde{\mathcal{C}} = \{\mathcal{C}_1, \mathcal{C}_2, \ldots, \mathcal{C}_m\}$, where each $\mathcal{C}_i \subset V$ denotes a cluster of nodes. The simplified network should retain key nodes and connections, preserving the core community structure of the original network. The proposed sparsification process is as follows (see Algorithm \ref{alg.graph_coarsening} for the pseudocode): We begin by determining the number of selected nodes from each community, that is

\begin{equation}
|\mathcal{C}'_i| = \left\lceil \frac{|\mathcal{C}_i|}{|V|} \times N_V^* \right\rceil ,
\label{node_size}
\end{equation}
where \( N_V^* \) denotes the target node number of the simplified network.

Once the proportional selection is established, the selection is guided by betweenness centrality, where nodes with higher centrality are prioritized, i.e.,

\begin{equation}
V' = \bigcup_{i=1}^{m} \{ v_j \in \mathcal{C}_i \mid b(v_j) \geq b(v_l), \forall v_l \in \mathcal{C}_i \setminus V'\},
\end{equation}
where $b(v)$ refers to the betweenness value of node $v$ and $|V' \cap \mathcal{C}_i| = |\mathcal{C}'_i|$.

Following node selection, a subgraph is induced by retaining only the edges between the selected nodes:
\begin{equation}
E' = \{ (u, v) \in E \mid u, v \in V' \}.
\end{equation}

To control the network's density, an edge-pruning step is applied if the number of edges in \( E' \) exceeds a predefined threshold \( |N^*_E| \). In this case, a subset of \( E' \) is randomly or strategically selected such that the total number of edges equals \( |N^*_E| \), and the rest are discarded.

In addition, isolated nodes will also be removed, and only the largest connected component is retained. The node and edge sets of refined network are updated as:
\begin{equation}
\begin{aligned}
V' &= \{ v \in V' \mid d(v) > 0 \}, \\
E' &= \{ (u, v) \in E' \mid u, v \in V' \}.
\end{aligned}
\end{equation}


\begin{algorithm}
\caption{Graph Sparsification}
\hspace*{0.02in} {\bf Input:} 
Graph $G = (V, E)$, the target node number and edge number of simplified network $N_V^*$ and $N_E^*$;\\
\hspace*{0.02in} {\bf Output:} Simplified network $G_s = (V', E')$
\begin{algorithmic}[1]
\State Initialize $V' \gets \emptyset$, $E' \gets \emptyset$
\State Identify the community structure \( \widetilde{\mathcal{C}} = \{\mathcal{C}_1, \mathcal{C}_2, \ldots, \mathcal{C}_m\} \)
\For{each community $\mathcal{C}_i$ in $\widetilde{\mathcal{C}}$}
\State Compute betweenness centrality $b(v)$ for each $v \in \mathcal{C}_i$
\State Sort nodes in $\mathcal{C}_i$ based on $b(v)$ in descending order
\State Select top $|\mathcal{C}'_i|$ nodes from $\mathcal{C}_i$ to include in $V'$
\EndFor
\State Construct $V' \gets \bigcup_{i=1}^m V'_i$
\State $E' \gets \{(u, v) \in E : u \in V' \text{ and } v \in V'\}$
\If{the number of edges in $E'$ exceeds a threshold $N_E^*$}
\State Randomly remove excess edges from $E'$ to meet the threshold
\EndIf
\State Remove isolated nodes from $V'$ 
\State Identify the largest connected component and update $G_s$
\State \Return $G_s$
\end{algorithmic}
\label{alg.graph_coarsening}
\end{algorithm}

\subsection{Community Merging}
Most existing community detection techniques often lead to clusters with highly imbalanced sizes, including many small or even single-node communities that contribute little to the overall network structure. This imbalance can undermine the effectiveness of the aforementioned graph sparsification, indicating the importance of merging these small-scale communities before graph sparsification. Note that the community merging step is unnecessary if the community detection algorithm already produces a balanced distribution. Given a graph \( G = (V, E) \) with an initial community structure \( \mathcal{C} \), where each community \( \mathcal{C}_i \subseteq \widetilde{\mathcal{C}} \), our goal is to reduce the total number of communities to a target number \( N_{\mathcal{C}}^* \). We define the size of each community \( \mathcal{C}_i \) as \( |\mathcal{C}_i| \), and the algorithm proceeds while the number of updated communities \( |\widetilde{\mathcal{C'}}| \) exceeds \( N_{\mathcal{C}}^* \). The community merging process involves several key steps:

\begin{enumerate}
    \item \textbf{Size Identification:} Identify the smallest community \( \mathcal{C}_{\text{min}} \), where \( |\mathcal{C}_{\text{min}}| = \min\{ |\mathcal{C}_i| \} \) for all \( i \in [1,m] \).
    
    \item \textbf{Connectivity Analysis:} For each edge \( (v_i, v_j) \in E \), increment an edge count between communities for edges where \( v_i \in \mathcal{C}_{\text{min}} \) and \( v_j \in \mathcal{C}_k \in \widetilde{\mathcal{C'}} \), and vice versa. This edge count \( e(\mathcal{C}_{\text{min}}, \mathcal{C}_k) \) measures the connectivity between \( \mathcal{C}_{\text{min}} \) and the community $\mathcal{C}_k$.
    
    \item \textbf{Determination of Closest Community:} Identify the community \( \mathcal{C}_{\text{close}} \) with the maximum edge count to \( \mathcal{C}_{\text{min}} \), i.e., 
    \[
    \mathcal{C}_{\text{close}} = \arg \max_{\mathcal{C}_k \in \widetilde{\mathcal{C'}}} e(\mathcal{C}_{\text{min}}, \mathcal{C}_k).
    \]
\end{enumerate}

After each merging operation, the community structure \( \widetilde{\mathcal{C'}} \) is updated, and the process repeats until the number of communities equals \( N_{\mathcal{C}}^* \). The process of community merging can be found in Algorithm \ref{alg:community_merging}.

\begin{algorithm}
\caption{Community Merging}
\label{alg:community_merging}
\begin{algorithmic}[1]
\Require Graph $G=(V,E)$, initial community structure $\widetilde{\mathcal{C}} = \{\mathcal{C}_1, \mathcal{C}_2, \ldots, \mathcal{C}_m\}$, target number of communities $N_{\mathcal{C}}^*$
\Ensure Updated community structure $\widetilde{\mathcal{C'}}$ with $|\widetilde{\mathcal{C'}}| = N_{\mathcal{C}}^*$
\While{$|\widetilde{\mathcal{C'}}| > N_{\mathcal{C}}^*$}
    \State Identify the smallest community $\mathcal{C}_{\min}$
    \For{each community $\mathcal{C}_k \in \widetilde{\mathcal{C'}}$ such that $\mathcal{C}_k \neq \mathcal{C}_{\min}$}
        \State Initialize $e(\mathcal{C}_{\min}, \mathcal{C}_k) \leftarrow 0$
        \For{each edge $(v_i,v_j) \in E$}
            \If{$(v_i \in \mathcal{C}_{\min}$ and $v_j \in \mathcal{C}_k)$ \textbf{or} $(v_i \in \mathcal{C}_k$ and $v_j \in \mathcal{C}_{\min})$}
                \State $e(\mathcal{C}_{\min}, \mathcal{C}_k) \leftarrow e(\mathcal{C}_{\min}, \mathcal{C}_k) + 1$
            \EndIf
        \EndFor
    \EndFor
    \State  Identify $\mathcal{C}_{\text{close}}$ to $\mathcal{C}_{\min}$   and merge them
    \State Update the community structure $\widetilde{\mathcal{C'}}$
\EndWhile
\State \Return $\widetilde{\mathcal{C'}}$
\end{algorithmic}
\end{algorithm}

\subsection{MLLM-based Evolutionary Optimizer}
When solving graph-structured combinatorial problems, traditional EAs typically utilize list-based encodings, consisting of sequences of node indices. While this approach is straightforward, it has some inherent drawbacks. Crossover and mutation operations could be compromised due to the absence of topological awareness concerning the network's structure. 
\begin{algorithm}
\caption{Visual Evolutionary Optimization}
\hspace*{0.02in} {\bf Input:} 
Graph $G = (V, E)$, thresholds $N_V^*$ and $N_E^*$\\
\hspace*{0.02in} {\bf Output:} 
Optimal node set $\mathcal{S'}$
\begin{algorithmic}[1]
    \If{$|V| > N_V^*$ \textbf{and} $|E| > N_E^*$}
        \State $G_s \gets$ Sparsify($G, \theta$)
        \State $G \gets G_s$
    \EndIf
    \State $\mathcal{P} \gets$ Initialize population with MLLMs by encoding $G$ into images
    \While{termination criteria not met}
        \State $\text{Fitness} \gets$ Evaluate the fitness of each node set in $\mathcal{P}$
        \State $\text{Parents} \gets$ Select individuals from $\mathcal{P}$ based on fitness for reproduction
        \State $\text{Offspring} \gets$ Generate new solutions by applying MLLM-based crossover to images of parent solutions
        \State Apply MLLM-based mutation to images of offspring solution to alter node sets
        \State $\mathcal{P} \gets$ Update the population with new offspring
    \EndWhile
    \State $\mathcal{S'} \gets$ Select the solution with the highest fitness in $\mathcal{P}$
    \State \Return $\mathcal{S'}$
\end{algorithmic}
\label{alg:VEO}
\end{algorithm}

\begin{figure*}[htbp]
\centering
\includegraphics[height=10cm,width=18cm]{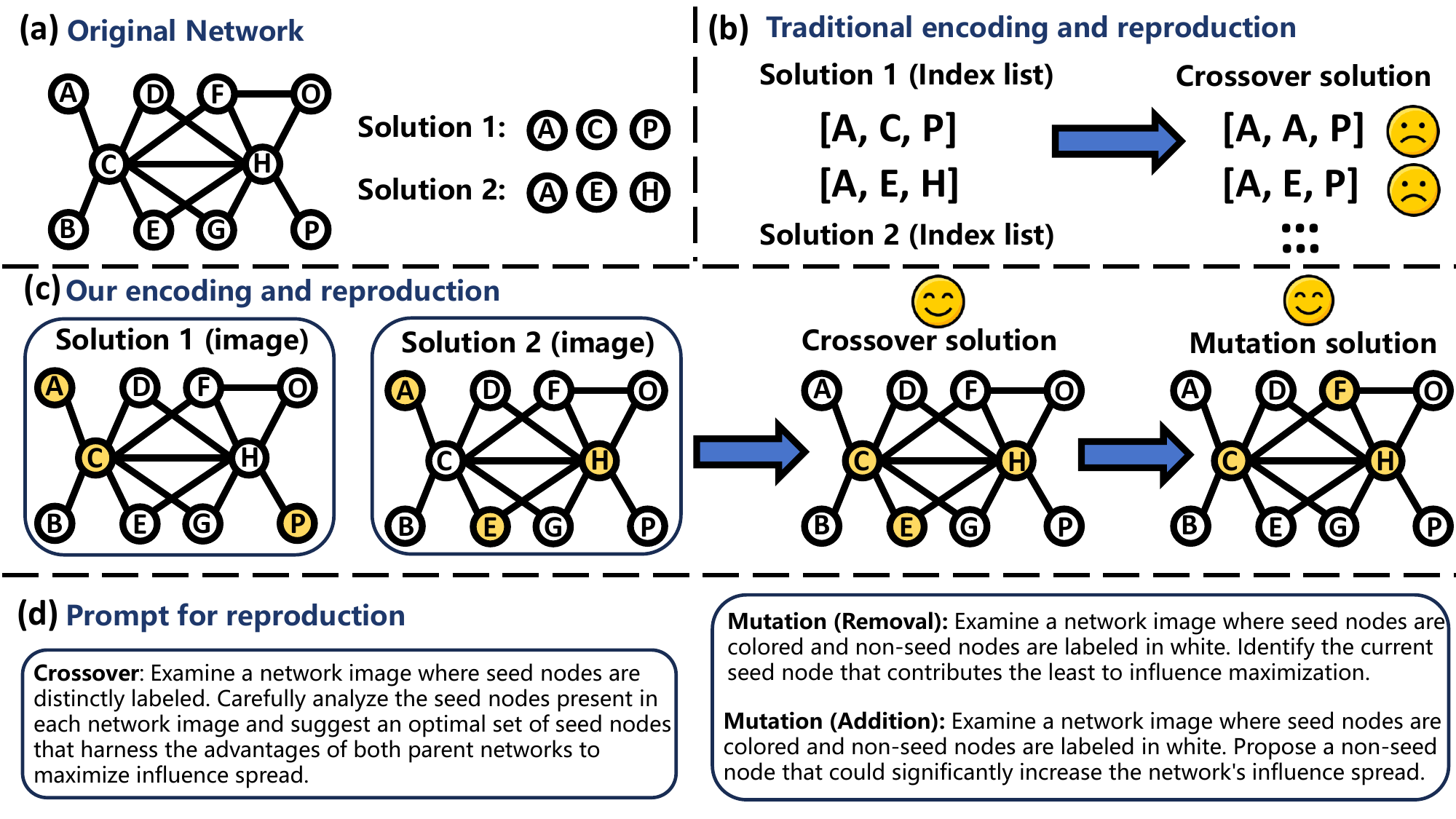}
\caption{The diagram of MLLM-driven VEO and the comparison with traditional evolutionary optimization, illustrated with influence maximization. (a): The original network presents a graph with nodes and edges, where two solutions (seed sets) are shown: Solution 1 [A, C, P] and Solution 2 [A, E, H], representing selected nodes to maximize influence spread. (b): Traditional encoding and reproduction represent solutions as index lists. However, the crossover process often generates invalid solutions, such as duplicate nodes (e.g., [A, A, P]) or suboptimal combinations (e.g., [A, E, P] where nodes A and E are low-degree). (c): Our scheme encodes the solution in the form of an image where seed nodes of the solution are distinctly highlighted. Crossover intelligently combines nodes from parent solutions to produce valid offspring by filtering out low-degree nodes (e.g., nodes A and P), while mutation replaces ineffective nodes with more suitable ones to enhance influence spread (e.g., replacing node E with node F). This approach ensures validity and better solution quality. (d): Prompt for reproduction provides instructions for an MLLM: in crossover, the MLLM examines network images and suggests an optimal combination of seed nodes; in mutation, it identifies ineffective nodes for replacement by removing nodes with minimal influence and proposing better non-seed nodes. }
\label{diagram}
\end{figure*}

\begin{table*}[h!]
\centering
\caption{Structure of prompts for MLLM-based evolutionary operators of different phases. The prompt begins with context-setting, which introduces the input information and clarifies the role of the agents. The latter part of the prompt specifies the desired output format and includes necessary guidance or restrictions.}
\label{table:prompts}
\begin{tabular}{m{2.8cm}m{7.5cm}m{6cm}}
\Xhline{5\arrayrulewidth}
\textbf{Task} & \textbf{Context-setting prompt} & \textbf{Output directive prompt} \\ \hline
\textbf{Initialization (Intelligent Selector)} & You are an expert in network science and will be provided with one network in the form of an image. Please help me intelligently select nodes as the diffusion seeds in this network to achieve influence maximization. & Only provide a list of node indices separated by commas. \\ \hline
\textbf{Initialization (High-degree Spread Selector)} & You are an expert in network science and will be provided with one network in the form of an image. Please help me intelligently select nodes as the diffusion seeds in this network to achieve influence maximization. & Here are some tips: (1) Choose large-betweenness nodes. (2) Pick nodes spread across different center parts of the network. Only provide a list of node indices separated by commas. \\ 
\hline
\textbf{Initialization (High-degree Central Selector)} & You are an expert in network science and will be provided with one network in the form of an image. Please help me intelligently select nodes as the diffusion seeds in this network to achieve influence maximization. & Here are some tips: (1) Choose large-degree nodes. (2) Pick nodes at the center place of the network. Only provide a list of node indices separated by commas. \\ 
\hline
\textbf{Crossover} & Examine a network image where seed nodes are distinctly labeled. Carefully analyze the seed nodes present in each network image and suggest an optimal set of seed nodes that harness the advantages of both parent networks to maximize influence spread. & Focus on selecting high-degree nodes or nodes in strategic positions that significantly enhance network connectivity. Provide your answer as a list of node indices, separated by commas. \\ 
\hline
\textbf{Mutation \quad \quad \quad (Two-Phase, Removal)} & Examine a network image where seed nodes are colored and non-seed nodes are labeled in white. Identify the current seed node that contributes the least to influence maximization. & Focus on nodes that appear trivial or less connected. Provide the index of this node. \\ 
\hline
\textbf{Mutation \quad \quad \quad (Two-Phase, Addition)} & Examine a network image where seed nodes are colored and non-seed nodes are labeled in white. Propose a non-seed node that could significantly increase the network's influence spread. & Focus on nodes with higher degrees or strategically critical positions in the network. Provide the index of this node.\\ 
\hline
{\textbf{Mutation \quad \quad \quad (One-shot)}} & {Examine a network image where seed nodes are colored and non-seed nodes are labeled in white. Identify the current seed node that contributes the least to influence maximization and propose a non-seed node in white that could significantly increase the network's influence spread.} &  {Provide the answer as a list: the first element is the index of the seed node to remove, and the second element is the index of the non-seed node to add.}\\ 
\Xhline{5\arrayrulewidth}
\end{tabular}
\end{table*}

To overcome these challenges, we propose a new context-aware framework VEO that incorporates visual encoding and direct interaction with network images by utilizing the capabilities of MLLMs. This method transforms the solution into an image format, enhancing machines' understanding of how specific nodes influence particular tasks. The diagram of our method is shown in Figure \ref{diagram} and the process of MLLM-based evolutionary optimization is in Algorithm \ref{alg:VEO}.

To ensure scalability for large networks, we first address cases where the network size exceeds predetermined thresholds to ensure the plotted network is manageable to MLLMs. Formally,
\begin{equation}
G = 
\begin{cases} 
\mathcal{S}(G; \theta), & \text{if } |V| > N_V^* \text{ and } |E| > N_E^*, \\[1mm]
G, & \text{otherwise.}
\end{cases}    
\end{equation}

To facilitate visual integration into our evolutionary framework, we define two distinct functions for generating image representations, one for initializing the population and another for encoding candidate solutions for crossover and mutation. First, we define the visualization function for initialization as
\begin{equation}
\mathcal{I}_{\text{init}}(G, \Theta) = \mathbf{Visualize}(G, \Theta),
\end{equation}
which produces an image of the graph \(G\) using the layout \(\Theta\) via visualization software. In contrast, to encode specific candidate solutions, we define the function
\begin{equation}
\mathcal{I}(G, \Theta, s) = \mathbf{Visualize}(G, \Theta, s),
\end{equation}
which generates an image by plotting the graph \(G\) and visually highlighting the nodes of the solution \(s\) with the layout \(\Theta\).

Based on this image-encoding representation, we can define the following MLLM-based evolutionary operator and the full prompt can be found in Table \ref{table:prompts}.

\begin{definition}\textbf{MLLM-based Initialization:}
Given a graph \(G\), and an initialization strategy \(\alpha\) (which can focus on different aspects) and layout style \(\Theta\), the initial population \(\mathcal{P}\) is generated as:
\begin{equation}
\mathcal{P} = \textbf{MLLM\_init}(\mathcal{I}_{\text{init}}(G, \Theta), \alpha; N_p) = \{ s_1, s_2, \ldots, s_{N_P} \},    
\end{equation}
where \(N_P\) denotes the population size.
\end{definition}

\begin{definition} \textbf{MLLM-based Crossover:}
For two parent solutions $s_i, s_j \in \mathcal{P}$, the crossover operator guided by MLLM generates an offspring \(s'\) as:
\begin{equation}
s' = {\textbf{MLLM\_crossover}}\big(\mathcal{I}(G, \Theta, s_i), \mathcal{I}(G, \Theta, s_j); \mathbb{P}_c\big),
\end{equation}
where \(\mathbb{P}_c\) is the crossover rate.
\end{definition}

\begin{definition}\textbf{MLLM-based Mutation:}
A mutation operator is similarly applied on the network, yielding:
\begin{equation}
s'' = \textbf{MLLM\_mutate}\big(\mathcal{I}(G, \Theta, s'); \mathbb{P}_m\big),
\end{equation}
where \(\mathbb{P}_m\) is the mutation rate.
\end{definition}

\subsection{Complexity Analysis}
The complexity of the graph sparsification method centers on several key computations. The betweenness centrality is calculated with a complexity of \(O(|V||E|)\). Constructing the subgraph and adjusting edges are achieved in \(O(|V'| + |E'|)\). The process of community merging iterates $(|\widetilde{\mathcal{C}}| - |\widetilde{\mathcal{C'}}|)$ times, where \( |\widetilde{\mathcal{C}}| \) represents the initial number of communities and \( |\widetilde{\mathcal{C'}}| \) the target count. Each iteration involves identifying the smallest community and calculating its edge connections, requiring \( O(|E|) \) time. The running time of MLLM-based evolutionary operators depends on the model’s architecture, with lighter and more efficiently optimized models achieving quicker results.

\section{Experimental studies}\label{sec.experiment}
In the experiment, we examine the effectiveness of MLLM in the evolutionary stages, as well as the parameter sensitivity.

\subsection{Experimental Settings and Dataset}
In the validation, we evaluate our VEO on eight real-world networks obtained from the Network Repository (\textit{https://networkrepository.com/}). The number of nodes and edges in the simplified networks is fixed at 50 and 100, respectively, to remain manageable for MLLMs. The probability of crossover and mutation are set to 0.2 and 0.1. The size of the initialized population is set to 15 and each strategy is prompted to generate 5 individuals. The reported result is averaged from 20 independent simulations. Communities with fewer than 2\% of the total nodes will be merged into their nearest, larger community. The initial community distribution of networks is obtained by the FastGreedy algorithm \cite{clauset2004finding}. The number of selected seeds is set to 5 for Netscience and USAir and is set to 10 for the rest of the larger-scale network. The backbone MLLM is \textit{Gpt-4o-2024-11-20}. {The visualization settings for different evolutionary operators are listed in Table \ref{tab:vis_settings}.}

\begin{table}
\centering
\caption{Basic network information including the number of nodes $|\text{V}|$, the number of edges $|\text{E}|$, {the average degree $\langle \text{K}\rangle$, the clustering coefficient $\text{CC}$, and the average shortest path length $\langle \text{d}\rangle$.} $|{\mathcal{C}}|$ and $|{\mathcal{C'}}|$ denote the number of communities in the original and simplified networks.}
\resizebox{0.49\textwidth}{!}{
\begin{tabular}{lccccccccc}
\Xhline{5\arrayrulewidth}
\textbf{Network} & 
$\mathbf{|V|}$ & 
$\mathbf{|E|}$ & 
$\mathbf{\langle K\rangle}$ & 
$\mathbf{CC}$ & 
$\mathbf{\langle d\rangle}$ & 
$\mathbf{|{\mathcal{C}}|}$ & 
$\mathbf{|{\mathcal{C'}}|}$ \\
\hline
\textbf{USAir} & 332 & 2,126&12.81&0.40&2.73 & 7& 3 \\
\textbf{Netscience} & 379 & 914& 4.82  & 0.74 & 6.06 & 19 & 4\\
\textbf{Polblogs} & 1,222 & 16,717&27.35&0.23&2.74 & 10 & 2 \\
\textbf{Facebook} & 4,039 & 88,234& 43.69 & 0.61 & 3.69 & 13 & 8 \\
\textbf{WikiVote} & 7,066 & 100,736& 28.51 &  0.21 & 3.25 & 31 & 3 \\
\textbf{Rutgers89} & 24,568 & 784,596&63.87&0.13&3.10  & 57 & 4 \\
\textbf{MSU24} & 32,361 & 1,118,767&69.14&0.12&3.04  & 29 & 4 \\
\textbf{Texas84} & 36,365 & 1,590,651 &87.48&0.10&2.89 &  45 &4  \\
\Xhline{5\arrayrulewidth}
\label{data}
\end{tabular}}
\end{table}

\begin{table}[h!]
\caption{{Visualization settings summary.}}
\centering
\begin{tabular}{l|l}
\Xhline{5\arrayrulewidth}
{\textbf{Setting}} & {\textbf{Value}} \\
\hline
{\textbf{Graph Library}} & \texttt{plotly.graph\_objs} \\
\hline
{\textbf{Node Size}} & 35 \\
\hline
{\textbf{Label Size}} & 22 \\
\hline
{\textbf{Canvas Size}} & 1200 × 1200 px \\
\hline
{\textbf{Layout}} & Kamada-Kawai \\
\hline
{\textbf{Color}} &
\begin{tabular}[c]{@{}l@{}}
Init. \& Crossover: \texttt{\#2F7FC1}\\
Mutation: Solution\texttt{\#2F7FC1},\\ Non-solution\texttt{\#FFFFFF}
\end{tabular} \\
\hline
{\textbf{Node Labels}} &
\begin{tabular}[c]{@{}l@{}}
Init. \& Mutation: All nodes \\
Crossover: Solution only
\end{tabular} \\
\Xhline{5\arrayrulewidth}
\end{tabular}
\label{tab:vis_settings}
\end{table}

\subsection{Benchmark}
We validate the effectiveness of MLLMs in both of initialization and reproduction, thus the benchmark has two parts:

\subsubsection{Initialization}

\textbf{Random Initialization} randomly selects nodes from the entire set of nodes in the original graph to generate the initial population. \textbf{Refined Random Initialization} randomly selects nodes from the set of nodes in the simplified graph to generate the initial population. \textbf{High-Degree Initialization} and \textbf{High-Betweenness Initialization} generate population by randomly sampling nodes from a preselected set of high-betweenness and high-degree nodes, respectively.
\textbf{MLLM-based Initialization} utilizes MLLMs to intelligently generate candidate solutions. Three distinct agents, each implementing a unique strategy, are employed to effectively initialize the population (see Table \ref{table:prompts}).

\subsubsection{Reproduction} This stage involves crossover and mutation operations. \textbf{Normal evolutionary optimization} refers to the approach where solutions are encoded using indices, with crossover and mutation operations performed randomly. In contrast, \textbf{MLLM-based reproduction} employs the proposed image-based encoding method, where crossover and mutation are executed by MLLMs.



\subsection{Examination of MLLM as Initialization Operator}



\begin{table*}[]
\captionof{table}{Statistical analysis (Mean and Standard Deviation (SD)) of fitness values obtained by different population initialization modes across eight networks. The optimization follows the normal evolutionary optimization and the population evolves for 50 generations.}
\centering
\begin{tabular}{lcccccccc}
\Xhline{5\arrayrulewidth}
\multirow{2}{*}{\textbf{Networks}} & \multicolumn{5}{c}{\textbf{Initialization modes}}   \\
\cmidrule(lr){2-6} 
& \textbf{Random}    & \textbf{Refined Random}    & \textbf{High-Degree}    & \textbf{High-Betweenness}    & \textbf{MLLMs}       \\
\midrule
\textbf{USAir} &48.84$\pm$2.23(-) & 49.27$\pm$1.57(-) & 49.12$\pm$1.88($\approx$) & 49.19$\pm$1.51($\approx$)& \textbf{49.92$\pm$1.98}\\
\textbf{Netscience} &16.24$\pm$0.56(-) & 15.89$\pm$0.80(-) & 16.40$\pm$0.53($\approx$) & 16.35$\pm$0.41(-)& \textbf{16.63$\pm$0.26}\\
\textbf{Polblogs} &183.32$\pm$13.11(-) & 213.77$\pm$5.94($\approx$) & 203.93$\pm$8.09(-) & 206.54$\pm$8.04(-)& \textbf{216.57$\pm$7.43}\\
\textbf{Facebook} &236.46$\pm$49.69(-) & 407.45$\pm$8.86($\approx$) & 380.77$\pm$31.13(-) & 376.12$\pm$36.48(-)& \textbf{410.57$\pm$5.80}\\
\textbf{WikiVote} &315.58$\pm$28.19(-) & 545.13$\pm$15.67(-) & 498.45$\pm$25.11(-) & 485.01$\pm$22.03(-)& \textbf{563.67$\pm$10.94}\\
\textbf{Rutgers89} &327.80$\pm$72.36(-) & 778.96$\pm$25.65(-) & 692.16$\pm$44.54(-) & 671.41$\pm$34.45(-)& \textbf{790.93$\pm$8.36}\\
\textbf{MSU24}&363.87$\pm$102.34(-) & 1152.17$\pm$27.81(-) & 1098.03$\pm$52.09(-) & 1033.17$\pm$113.48(-)& \textbf{1170.79$\pm$15.29}\\
\textbf{Texas84}&543.29$\pm$147.81(-) & 2594.51$\pm$144.18($\approx$) & 2063.10$\pm$208.67(-) & 2109.58$\pm$190.13(-)& \textbf{2629.62$\pm$80.04}&\\
\Xhline{5\arrayrulewidth}
{$\textbf{+}/\approx/\textbf{-}$} &{0/0/8}&{0/3/5}&{0/2/6}&{0/1/7}&-\cr
\hline
{\textbf{Avg ranking}} &{4.88}&{2.38}&{3.25}&{3.50}&\textbf{1.00}\cr
\Xhline{5\arrayrulewidth}
\end{tabular}
\label{sta_initialization}
\end{table*}

To compare different methods, we use the Wilcoxon test with a 95\% confidence interval to statistically examine their difference. The `+' indicates statistically better performance, `$\approx$' indicates no significant difference, and `-' indicates statistically worse performance. Table \ref{sta_initialization} presents the optimization results regarding fitness values achieved through different initialization modes across eight networks. The fitness values are reported as averages with standard deviations (SD). The MLLM-based method consistently outperforms other modes across all networks, as indicated by the highest fitness values and its average ranking of 1.00, demonstrating its superiority. This analysis highlights the superior capability of MLLMs in initializing the population for evolutionary optimization more effectively compared to traditional approaches. {In addition, the results show that Refined Random initialization (sampled from the simplified network) outperforms the other three non-MLLM initialization methods (derived from the original network) in 7 out of 8 cases except for Netscience, demonstrating the effectiveness of our graph sparsification approach.}

\begin{table}[ht]
\centering
\caption{The ANOVA results for different graphs analyzing significant differences among agents.}
\label{std.anova}
\begin{tabular}{lccc}
\Xhline{5\arrayrulewidth}
\textbf{Graph}   & \textbf{F-statistic} & \textbf{P-value}   & \textbf{Different} \\ \hline
\textbf{Netscience}       & 26.35              & $2.92 \text{E-}8$ & Yes                             \\ 
\textbf{USAir}            & 8.42               & $2.76 \text{E-}4$  & Yes                             \\ 
\textbf{Polblogs}         & 4.90               & $8.00 \text{E-}3$  & Yes                            \\ 
\textbf{Facebook}         & 3.56               & $2.96\text{E-}2$  & Yes                             \\ 
\textbf{WikiVote}            & 5.53               & $4.36 \text{E-}3$  & Yes                             \\ 
\textbf{Rutgers89}        & 3.30               & $3.80 \text{E-}2$  & Yes   \\    
\textbf{MSU24}            & 1.24              & $2.91 \text{E-}1$  & No                              \\ 
\textbf{Texas84}          & 2.03               & $1.33 \text{E-}1$  & No                              \\ 
\Xhline{5\arrayrulewidth}
\end{tabular}
\end{table}

In addition, we investigate the accuracy of MLLMs in following instructions. Specifically, we examine whether the initial population obtained through various strategies exhibits any differences. Table \ref{std.anova} presents the ANOVA results for three different agent-based initialization methods used in various networks. These agents operate under strategies formulated by MLLMs, which demonstrate an acute awareness of network structural topologies and can execute tasks according to specific instructions. The results indicate significant differences in the performance of these agents in 6 out of 8 networks examined, with only MSU24 and Texas84 showing no significant differences among the agents' strategies. This result also demonstrates the spatial intelligence of MLLMs and their potential in the graph problems.

\begin{figure*}[htbp]
\centering
\includegraphics[height=7.cm,width=15.5cm]{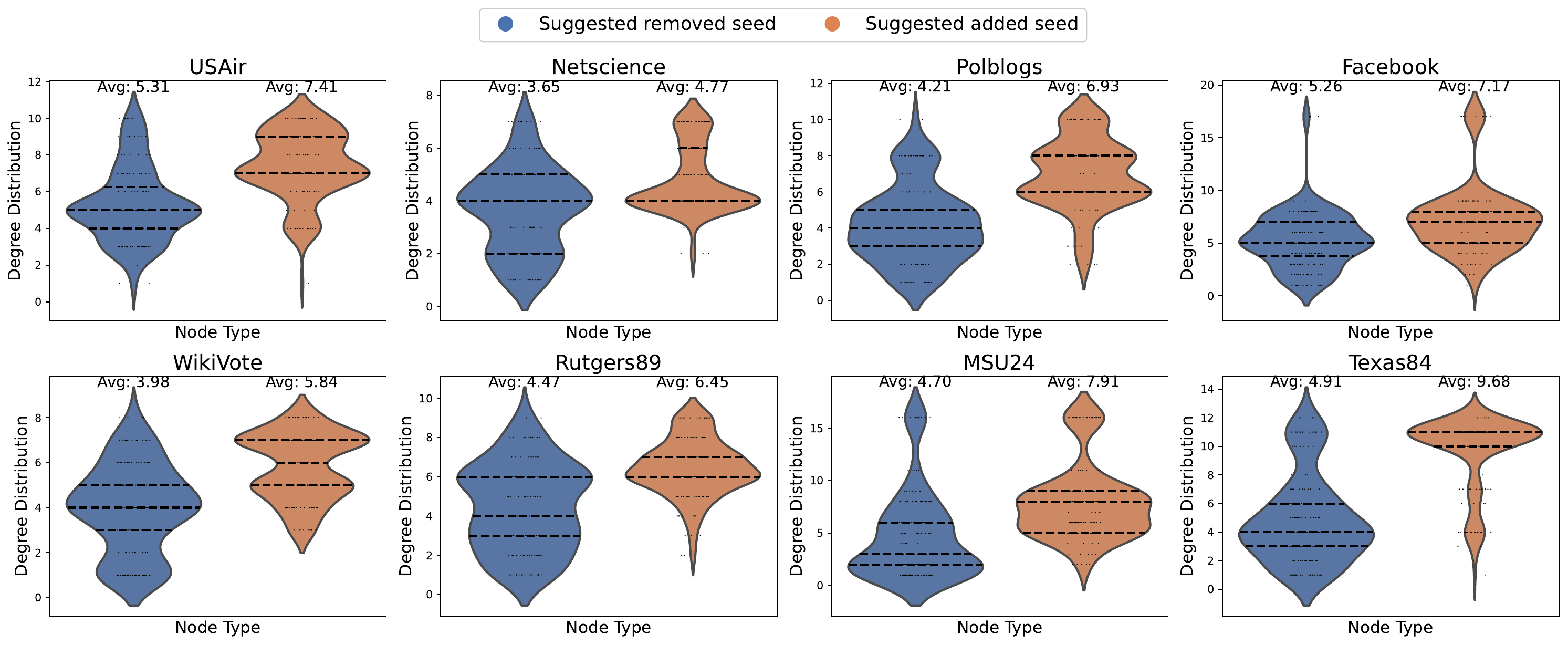}
\caption{The comparative degree analysis of node selection for addition and removal suggested by MLLMs during the mutation stage in different networks.}
\label{mutation_check}
\end{figure*}

To gain a perspective on the node selection behavior of MLLMs during mutation, we compared the degree distribution of nodes suggested for removal and addition by MLLMs in Figure \ref{mutation_check}. As seen, the average degree of nodes suggested for addition is clearly higher than those suggested for removal, indicating a strategy favoring the addition of more highly connected nodes to maximize influence or connectivity potentially. The marked contrast in the degree distributions indicates the MLLMs' awareness of the network structure and the ability to intelligently identify critical nodes, thereby enhancing the overall strategy for influence maximization.

\begin{table*}[]
\captionof{table}{{Statistical analysis (Mean and Standard Deviation (SD)) of fitness values obtained by different reproduction modes across eight networks, using MLLM-driven evolutionary operations. The networks are visualized in KK and FR styles, and the MLLM operates under One-Shot and Two-Phase mutation strategies. The population is initialized using the refined random strategy within the sparsified network domain and then evolves for 10 generations.}}
\centering
\begin{tabular}{lcccccccc}
\Xhline{5\arrayrulewidth}
\multirow{2}{*}{\textbf{Networks}}  & \multicolumn{4}{c}{\textbf{Reproduction modes}}  \\
\cmidrule(lr){2-5} 
 & \textbf{Normal} & {\textbf{MLLM (One-Shot, KK)}} & {\textbf{MLLM (Two-Phases, FR)}} & \textbf{MLLM (Two-Phases, KK)}   \\
\midrule
\textbf{USAir} &44.03$\pm$3.17($\approx$) &43.46$\pm$2.62(-)  & 43.74$\pm$2.91(-) &\textbf{45.02$\pm$2.49}\\
\textbf{Netscience} &14.04$\pm$1.11(-)&14.55$\pm$1.15 ($\approx$)&\textbf{14.75$\pm$0.94}($\approx$)&{14.63$\pm$1.13}\\
\textbf{Polblogs} & 187.95$\pm$8.90($\approx$) &\textbf{193.65$\pm$10.50}($\approx$)&{191.44$\pm$8.82}($\approx$)&{190.75$\pm$6.94}\\
\textbf{Facebook} &354.58$\pm$36.72(-) &353.31$\pm$42.28($\approx$)&356.00$\pm$34.24($\approx$)&\textbf{367.90$\pm$27.03}\\
\textbf{WikiVote} &480.32$\pm$20.59(-) &487.36$\pm$24.42(-)&501.01$\pm$22.15($\approx$)&\textbf{506.40$\pm$22.91}\\
\textbf{Rutgers89}& 686.99$\pm$39.65(-) & 698.56$\pm$46.61($\approx$)&\textbf{718.94$\pm$44.73}($\approx$)&{710.39$\pm$45.16}\\
\textbf{MSU24}&1017.86$\pm$140.14(-) &1080.24$\pm$339.29($\approx$)&1071.14$\pm$39.23($\approx$)&\textbf{1084.14$\pm$39.20}\\
\textbf{Texas84}& 2197.38$\pm$225.31(-)&\textbf{2339.35$\pm$168.78}($\approx$)&2269.19$\pm$172.68($\approx$)&{2277.91$\pm$213.75}\\
\Xhline{5\arrayrulewidth}
\end{tabular}
\label{std_reproduction}
\end{table*}

\subsection{Ablation Study of MLLM as Reproduction Operator}

To demonstrate the effectiveness of VEO in reproduction (including crossover and mutation), Table \ref{std_reproduction} compares the fitness values achieved through `Normal' and `MLLMs' modes. {The results demonstrate that VEO (Two-Phases, KK) consistently outperforms conventional evolutionary optimization across networks of varying scales and complexities, highlighting the potential of MLLMs as effective evolutionary operators for graph-based combinatorial problems.}

{For generalizability, we intentionally employ a simple prompt strategy to test the inherent capability of MLLMs on such problems. The only strategy we adopt is splitting the mutation prompt into two steps, referred to as Two-Phases (see Table \ref{table:prompts}). We conduct an ablation study to compare this with a One-Shot prompting strategy. As shown in Table \ref{std_reproduction}, both One-Shot and Two-Phases approaches achieve improved fitness values, but the Two-Phases mutation strategy tends to slightly outperform the One-Shot method, suggesting that a clearer and stepwise prompt may better guide the model during evolution.}

{Furthermore, the results across networks plotted in both KK (Kamada–Kawai) and FR (Fruchterman–Reingold) styles indicate that the KK layout style generally achieves marginally better fitness results compared to those in the FR style. While most cases exhibit no statistically significant differences, the observed variations in certain instances indicate that the graph layout might be the critical factor affecting the performance of MLLM-driven optimization.}

\begin{figure*}[htbp]
\centering
\includegraphics[height=7.cm,width=15.5cm]{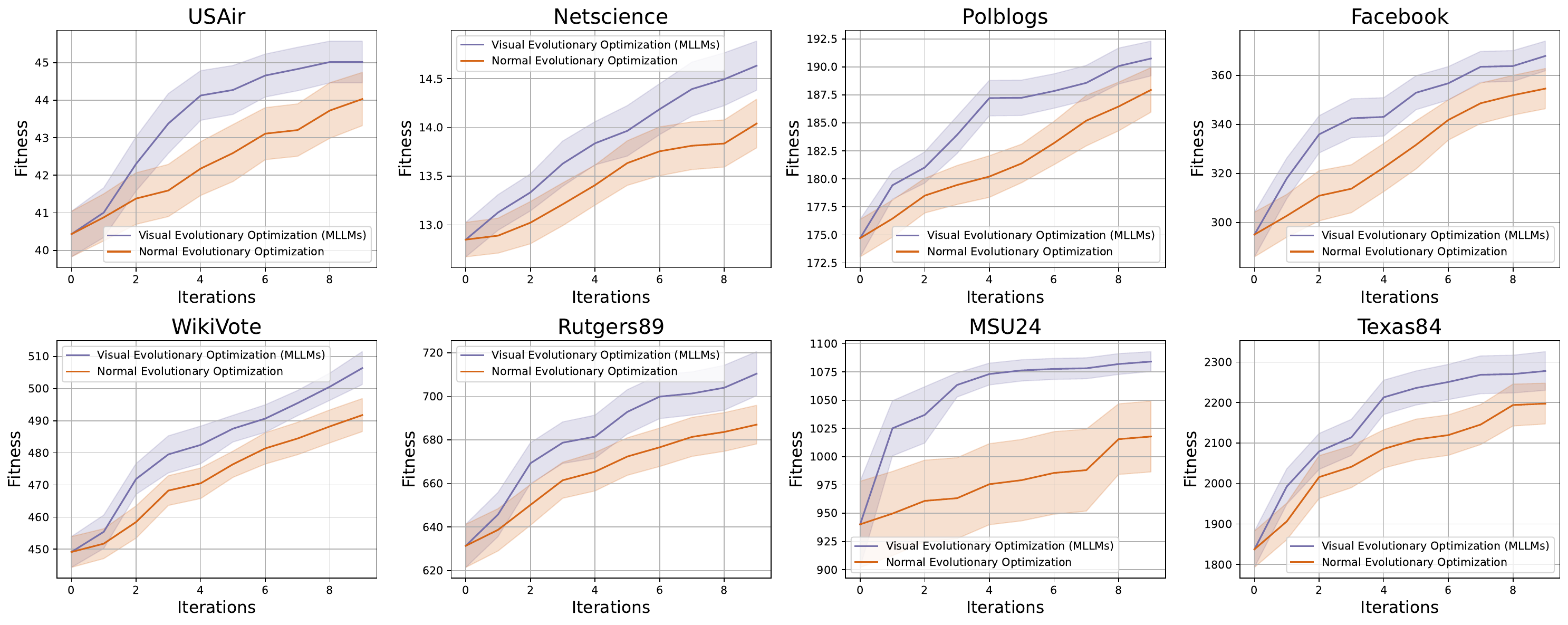}
\caption{Fitness evolution over iterations for different reproduction modes across eight different networks. Shaded areas indicate Standard Error of the Mean (SEM) over multiple runs.}
\label{reproduction_comparison}
\end{figure*}

Figure \ref{reproduction_comparison} presents the optimization landscape of VEO against traditional evolutionary optimization across eight networks. The VEO approach, facilitated by MLLMs, consistently outperforms the traditional method, achieving higher fitness scores and demonstrating accelerated improvements in the initial iterations. The shaded regions indicate variability in the results, showing the robustness and reliability of the VEO. These findings suggest that MLLMs provide insightful guidance, facilitating optimization in graph-structured combinatorial problems and achieving superior outcomes with fewer iterations.

\subsection{{Correctness and Reliability Examination}}

\begin{table*}[]
\captionof{table}{{Validation of output of MLLMs during different evolutionary stages. For initialization, tests include $\mathrm{T}_\mathrm{I}^1$: Valid Node Check (ensuring nodes belong to the valid set), $\mathrm{T}_\mathrm{I}^2$: Initialization Size Check (verifying solutions meet the required seed size), and $\mathrm{T}_\mathrm{I}^3$: Low-Degree Node Check (identifying nodes with low degrees). During crossover, validations include $\mathrm{T}_\mathrm{C}^1$: Crossover Size Check (ensuring offspring solutions meet size requirements), $\mathrm{T}_\mathrm{C}^2$: Duplicate Node Check (identifying duplicate nodes), and $\mathrm{T}_\mathrm{C}^3$: Parent Node Source Check (verifying nodes are derived from parent solutions). For mutation, tests consist of $\mathrm{T}_\mathrm{M}^1$: Node Presence Check (ensuring removed nodes exist in the solution), $\mathrm{T}_\mathrm{M}^2$: Mutation Valid Node Check (verifying added nodes belong to the valid set) and $\mathrm{T}_\mathrm{M}^3$: Mutation Repetitive Node Check (verifying added nodes already exist in the solution).}}
\centering
\resizebox{1\textwidth}{!}{
\begin{tabular}{lcccccccccccc}
\Xhline{5\arrayrulewidth}
\multirow{2}{*}{\textbf{Networks}} & \multicolumn{3}{c}{\textbf{Initialization (KK)}} & \multicolumn{3}{c}{\textbf{Initialization (FR)}} & \multicolumn{3}{c}{\textbf{Crossover}} & \multicolumn{3}{c}{\textbf{Mutation}} \\
\cmidrule(lr){2-4} \cmidrule(lr){5-7} \cmidrule(lr){8-10}\cmidrule(lr){11-13}
& $\mathbf{\mathrm{T}_\mathrm{I}^1}$ & $\mathbf{\mathrm{T}_\mathrm{I}^2}$ & $\mathbf{\mathrm{T}_\mathrm{I}^3}$ & $\mathbf{\mathrm{T}_\mathrm{I}^1}$ & $\mathbf{\mathrm{T}_\mathrm{I}^2}$ & $\mathbf{\mathrm{T}_\mathrm{I}^3}$ & $\mathbf{\mathrm{T}_\mathrm{C}^1}$ & $\mathbf{\mathrm{T}_\mathrm{C}^2}$ & $\mathbf{\mathrm{T}_\mathrm{C}^3}$ & $\mathbf{\mathrm{T}_\mathrm{M}^1}$ & $\mathbf{\mathrm{T}_\mathrm{M}^2}$& $\mathbf{\mathrm{T}_\mathrm{M}^3}$\\
\midrule
\textbf{USAir} &100.0\% & 100.0\% & 99.5\%  &100.0\% & 100.0\% & 99.9\%   & 100.0\%& 99.4\%& 100.0\%& 100.0\%& 100.0\%& 96.4\%\\
\textbf{Netscience} &100.0\% & 100.0\% & 98.7\%  &100.0\% & 100.0\% & 99.0\%   & 100.0\%& 100.0\%& 99.5\%& 99.5\% & 99.9\%& 89.2\%\\
\textbf{Polblogs} &100.0\% & 100.0\% & 99.6\%  &100.0\% & 100.0\% & 98.3\%  &100.0\% & 100.0\%  & 98.2\%&99.4\%& 100.0\%& 85.1\%\\
\textbf{Facebook} &100.0\% & 100.0\% & 98.7\%  &100.0\% & 95.0\% & 93.1\%  &99.5\% & 99.9\%  & 99.5\%& 100.0\%& 100.0\%& 76.3\%\\
\textbf{WikiVote} &100.0\% & 100.0\% & 99.5\% &100.0\% & 100.0\% & 96.4\%  &99.6\% & 100.0\%  & 100.0\%& 100.0\%& 100.0\%& 85.2\%\\
\textbf{Rutgers89} &100.0\% & 100.0\% & 97.6\%  &100.0\% & 100.0\% & 98.7\%   &99.9\% & 99.5\%  & 96.9\%& 99.9\%& 99.9\%& 86.5\%\\
\textbf{MSU24}&100.0\% & 100.0\% & 99.0\%  &100.0\% & 100.0\% & 97.6\%   &100.0\% & 99.5\%  & 97.4\%& 99.9\%& 99.9\%& 90.8\%\\
\textbf{Texas84} &100.0\% & 100.0\% & 99.0\%  &100.0\% & 100.0\% & 99.5\%  &100.0\% & 100.0\%  & 100.0\%& 98.8\%& 100.0\% &50.1\%\\
\Xhline{5\arrayrulewidth}
\end{tabular}}
\label{validation}
\end{table*}

Due to the commonly reported LLMs hallucination \cite{xu2024hallucination}, the validity examination is required to ensure the correctness. Table \ref{validation} reflects generally strong validation performance across most evolutionary stages, particularly during initialization and crossover. Tests $\mathrm{T}_\mathrm{I}^1$ and $\mathrm{T}_\mathrm{I}^2$ consistently achieve perfect scores across all networks, indicating robust mechanisms for validating node feasibility and ensuring correct solution sizes. Although the $\mathrm{T}_\mathrm{I}^3$ test, which checks for low-degree nodes, records slightly lower values, such as 93.1\% for Facebook in the FR method, it still demonstrates a high level of reliability, with minor room for refinement. Crossover validation remains solid, with nearly all networks achieving above 99\% in $\mathrm{T}_\mathrm{C}^1$ and $\mathrm{T}_\mathrm{C}^3$, and only slight drops in $\mathrm{T}_\mathrm{C}^2$, suggesting effective control over offspring structure. Mutation tests show more variability, with $\mathrm{T}_\mathrm{M}^3$ revealing opportunities for enhancement in managing repetitive node additions.  Overall, the result indicates the robust performance of MLLMs in maintaining high standards of the evolutionary process.

\begin{figure*}[htbp]
\centering
\includegraphics[height=7.5cm,width=15.5cm]{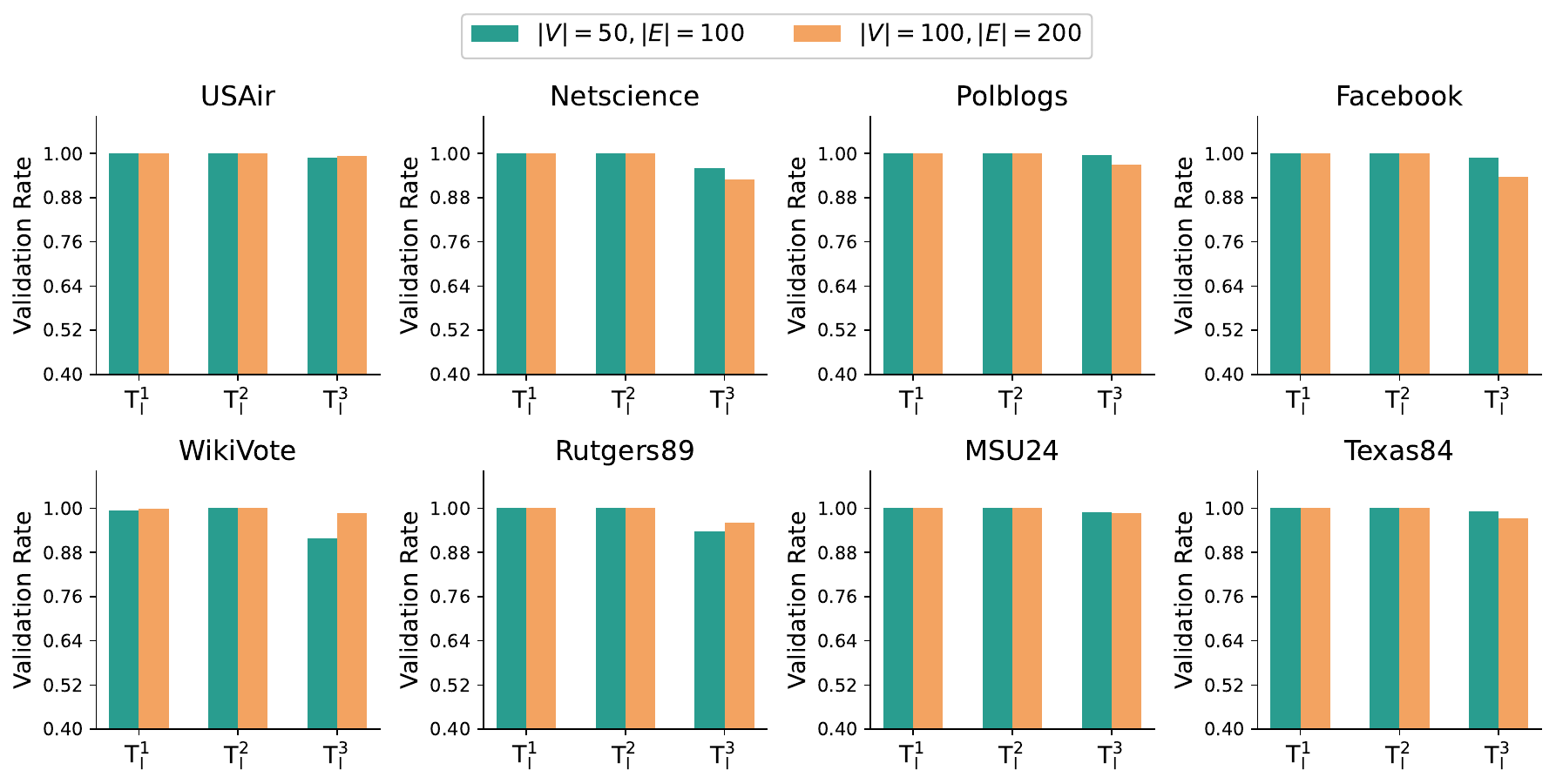}
\caption{{The comparative analysis of MLLMs output validity in initialization under different-size input graphs. Tests include $\mathrm{T}_\mathrm{I}^1$: Valid Node Check (ensuring nodes belong to the valid set), $\mathrm{T}_\mathrm{I}^2$: Initialization Size Check (verifying the required seed size is met), and $\mathrm{T}_\mathrm{I}^3$: Low-Degree Node Check (identifying low-degree nodes). The seed size is fixed at 10.}}
\label{factor_graph_size}
\end{figure*}

Furthermore, we examine the parameter sensitivity of MLLMs in terms of output validation. Figure \ref{factor_graph_size} presents a comparative analysis of MLLM output validity during initialization across networks of two different input sizes. Across all networks and tests, the validation rates remain consistently high, typically exceeding 95\%, which demonstrates the robustness of the initialization process. The $\mathrm{T}_\mathrm{I}^1$ and $\mathrm{T}_\mathrm{I}^2$ checks maintain near-perfect accuracy, showing that node validity and seed size constraints are reliably enforced regardless of graph size. The $\mathrm{T}_\mathrm{I}^3$ test, which identifies low-degree nodes, shows slightly more variation, particularly in networks like Facebook and WikiVote, where a moderate dip is observed in larger graphs. Despite this, all validation rates stay well above 88\%, reflecting the scalability and resilience of the MLLM framework, even as input complexity increases.

\begin{figure*}[htbp]
\centering
\includegraphics[height=7.5cm,width=15.5cm]{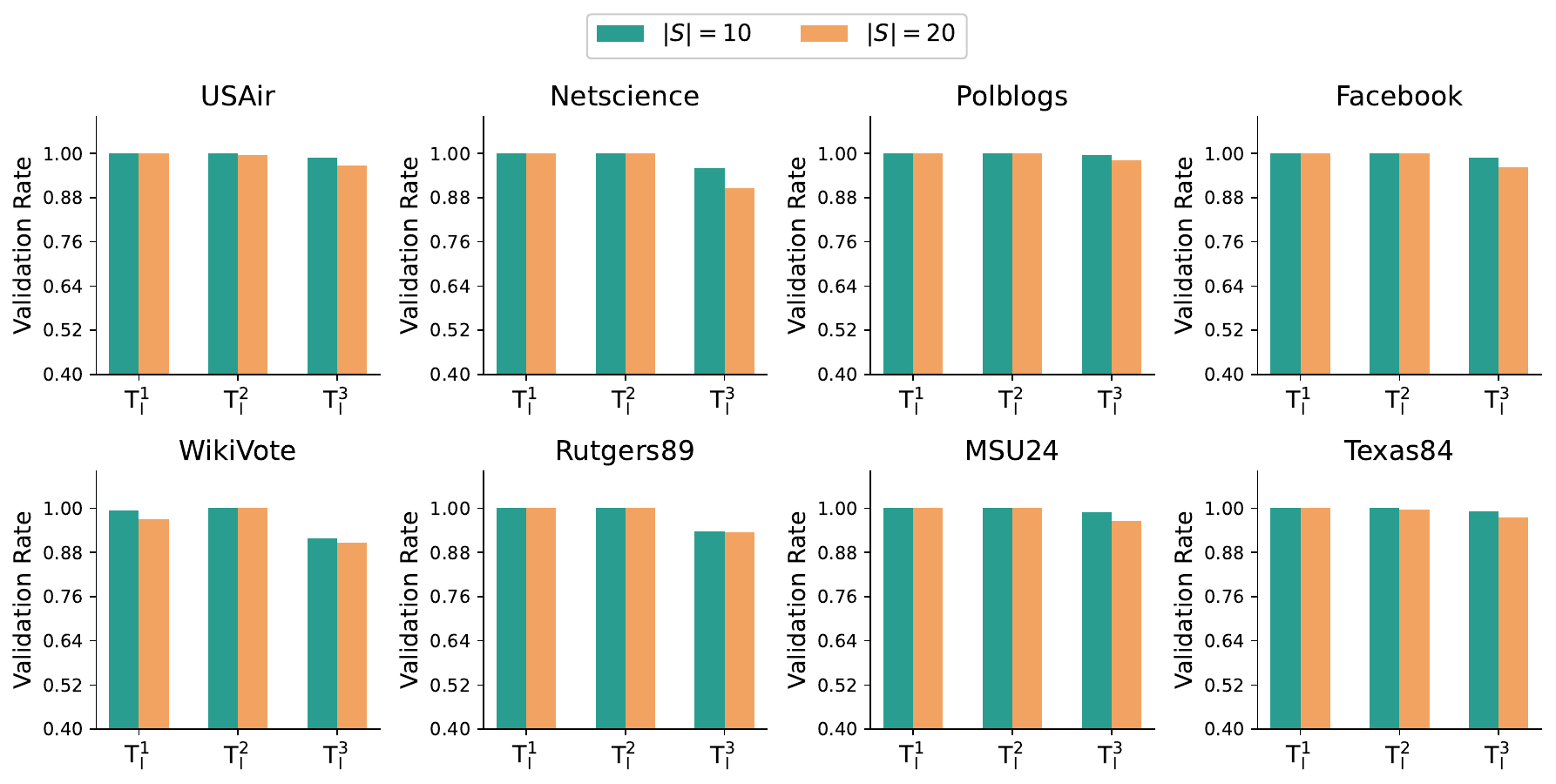}
\caption{The comparative analysis of MLLMs output validity in initialization under different seed size settings. Tests include $\mathrm{T}_\mathrm{I}^1$: Valid Node Check (ensuring nodes belong to the valid set), $\mathrm{T}_\mathrm{I}^2$: Initialization Size Check (verifying the required seed size is met), and $\mathrm{T}_\mathrm{I}^3$: Low-Degree Node Check (identifying low-degree nodes). The number of simplified graph is $\{|V|=50,|E|=100\}$.}
\label{factor_seed_size}
\end{figure*}

On the other hand, we also examine the influence of the required number of selected seeds on the validity, as shown in Figure \ref{factor_seed_size}. As observed, the validation rate for $\mathrm{T}_\mathrm{I}^1$ and $\mathrm{T}_\mathrm{I}^2$ are basically the same for $|S| = 10$ and $|S| = 20$ and there is a slight drop for $\mathrm{T}_\mathrm{I}^3$ when the required sample nodes increase. This result demonstrates the capability of MLLMs to deal with graph-related problems even when the complexity of tasks increases.

\subsection{Foundation Model Comparison}

\begin{figure*}[htbp]
\centering
\includegraphics[height=8cm,width=15.5cm]{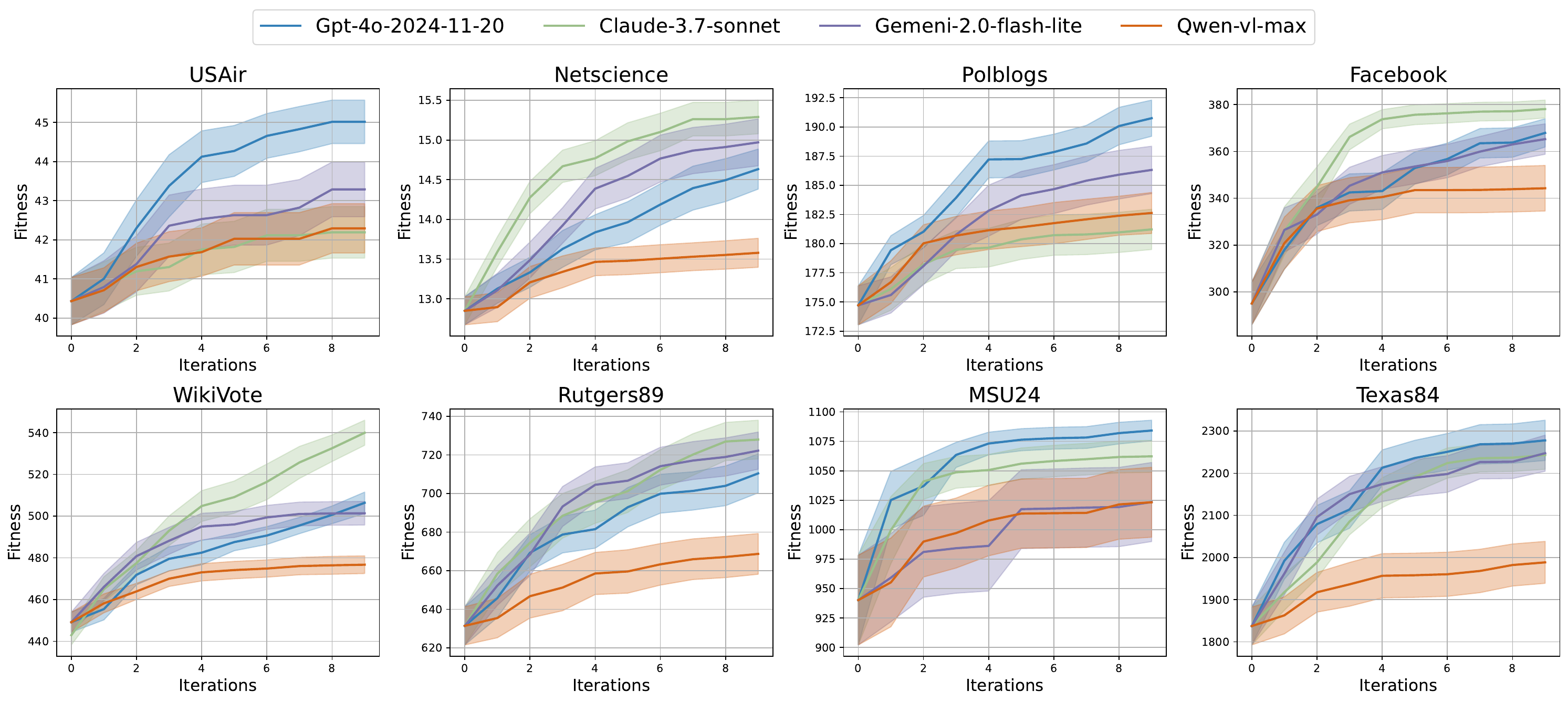}
\caption{{Fitness evolution over iterations for MLLM-based reproduction strategies using different models across eight different networks. Shaded areas indicate Standard Error of the Mean (SEM) over multiple runs.}}
\label{model_comparison}
\end{figure*}

{In terms of model selection, Figure \ref{model_comparison} presents the progression of fitness values over optimization iterations when different MLLMs are used for reproduction operations across various networks. Overall, all models show steady improvements in fitness as iterations proceed, confirming the effectiveness of VEO pipeline. Moreover, clear performance differences among models are observed. Gpt-4o-2024-11-20 and Claude-3.7-sonnet consistently achieve higher final fitness values across most networks, but the other models also demonstrated effective optimization trends, showing that the VEO framework is compatible with various MLLMs. Notably, the differences in performance suggest that the capabilities of each MLLM, such as visual understanding, layout comprehension, and generation quality, directly affect the optimization results.  These observations indicate that model selection is critical when using MLLM-based strategies for solving combinatorial problems.}

\begin{figure*}[htbp]
\centering
\includegraphics[height=7.2cm,width=15.5cm]{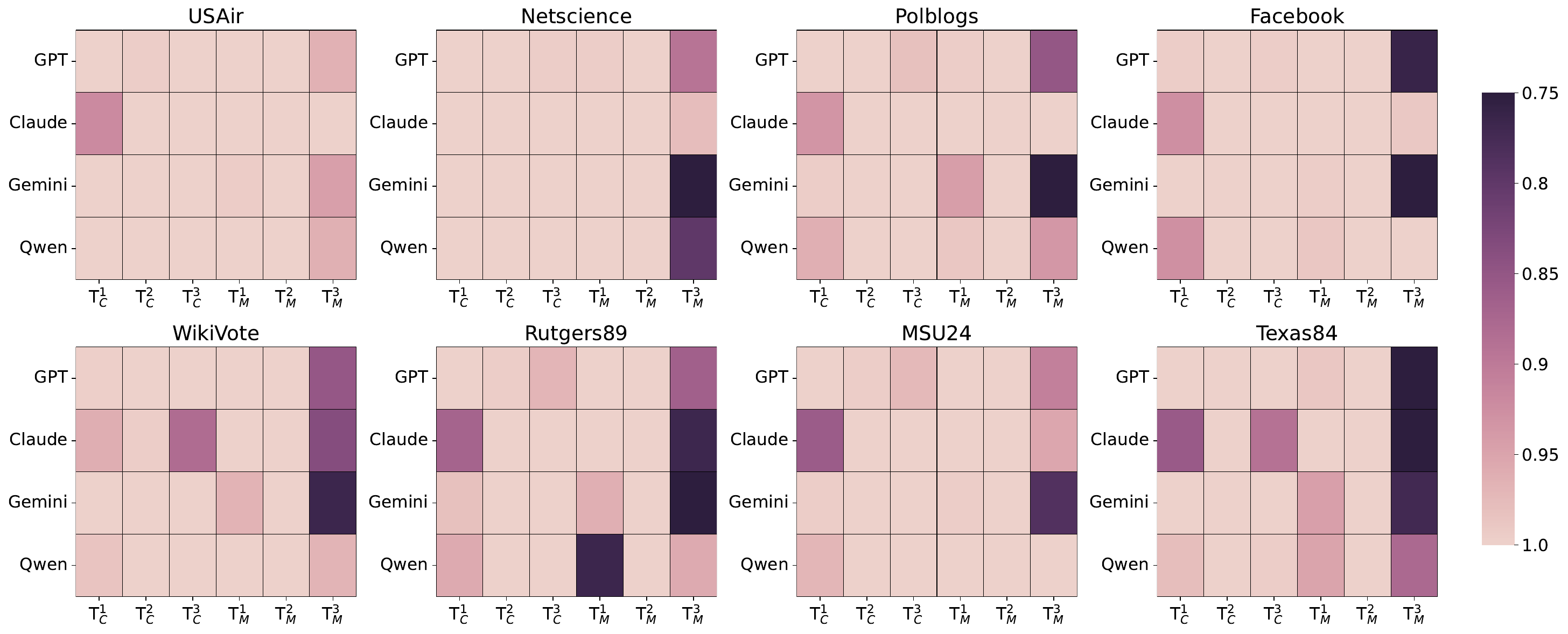}
\caption{{Validation rates for different MLLMs across various networks. Higher validation rates indicate greater reliability of reproduction operations, with darker colors corresponding to lower validation rates.}}
\label{model_validation}
\end{figure*}

{Figure \ref{model_validation} shows the validation rates of different MLLMs when performing reproduction. A higher validation rate indicates that the offspring generated by the model satisfy the problem constraints and maintain solution feasibility. While the reliability is subject to the specific model used, all models achieve strong validation rates, with nearly all results above 75\% and the majority exceeding 95\%. Additionally, mutation operations present slightly more difficulty for models compared to crossover. These results confirm that all evaluated MLLMs are reliable for reproduction in combinatorial optimization.}

\subsection{Computational Cost Analysis}

\begin{table}[]
\captionof{table}{{Average running time (in seconds) per API call for different MLLMs performing crossover (with 2 input images) and mutation (with 1 input image) during evolutionary optimization.}}
\centering
\begin{tabular}{lccccccccccc}
\Xhline{5\arrayrulewidth}
{\textbf{Model}}  &\textbf{Crossover}   & \textbf{Mutation}  \\
\midrule
\textbf{gpt-4o-2024-11-20} &2.46$\pm$0.52 & 1.75$\pm$0.41   \\
\textbf{Gemeni-2.0-flash-lite} &2.28$\pm$0.46 & 2.18$\pm$0.45  \\
\textbf{Claude-3.7-sonnet} &5.59$\pm$1.34 & 4.11$\pm$1.30  \\
\textbf{Qwen-vl-max}&3.11$\pm$0.95 & 1.81$\pm$0.56 \\
\Xhline{5\arrayrulewidth}
\end{tabular}
\label{time_comparison}
\end{table}

{To investigate the efficiency of VEO, we compare running time per API calls for crossover and mutation across different models in Table \ref{time_comparison}. The results clearly show that the running time is closely related to the choice of MLLM model, with Claude-3.7-sonnet being significantly slower than the others, while Gemini-2.0-flash-lite and Gpt-4o-2024-11-20 are much faster and more consistent. Additionally, the running time is influenced by the number of input images or input token numbers. Crossover operations (with 2 images) generally take longer than mutation (with 1 image) across all models.}

\begin{figure}[htbp]
\centering
\includegraphics[height=5cm,width=8.9cm]{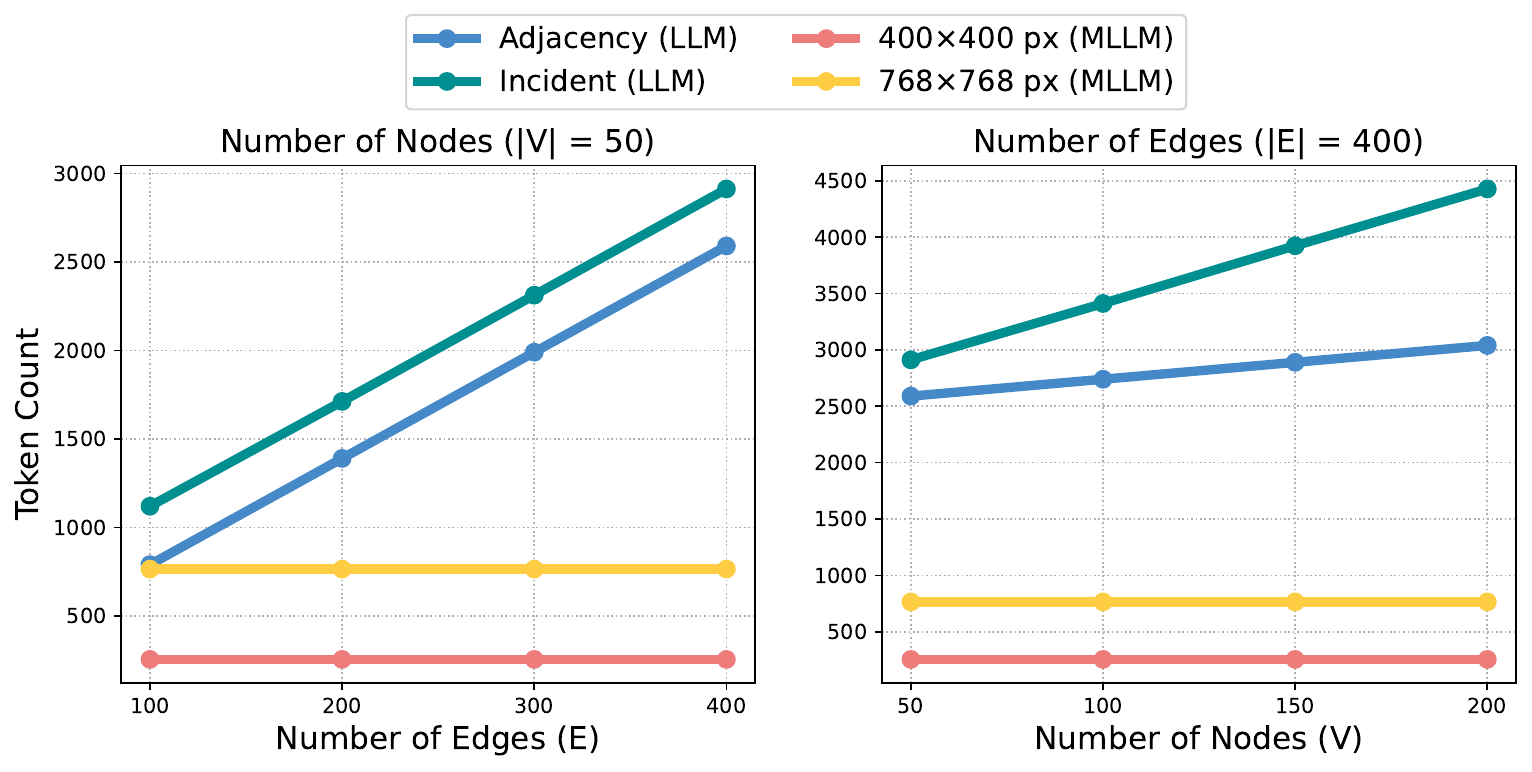}
\caption{{Token cost comparison for different input formats when depicting networks of varying sizes.}}
\label{token_comparison}
\end{figure}

{In our work, we focus on MLLM-based optimization of combinatorial problems, where a critical aspect is ensuring that the model effectively perceives and reasons over the underlying network structure. To achieve this, the majority of token usage is dedicated to depicting the network, either visually (via images) or textually to provide the LLMs or MLLMs with sufficient structural context. In contrast, the actual prompt instructions used for optimization are minimal and remain static across tasks, accounting for only a marginal portion of the overall token cost. Therefore, our cost analysis specifically isolates and compares the cost of network depiction in image and text formats, which represents the dominant share of token usage. Figure \ref{token_comparison} illustrates how the token cost scales with network size under different input representations for both LLM- and MLLM-based models. For LLM inputs (Adjacency and Incident methods) \cite{fatemi2023talk}, token counts grow significantly as the number of edges or nodes increases, highlighting a direct linear relationship between network size and input complexity. In contrast, for MLLM inputs based on images, the token cost remains almost constant regardless of the network scale, with differences between the 400×400 and 768×768 pixel image sizes. This indicates that visual representations via images offer much greater scalability for handling large networks, as they decouple the input size from the underlying network complexity. }

\subsection{{Generalizability Test}}

\begin{figure}[htbp]
\centering
\includegraphics[height=4cm,width=8.9cm]{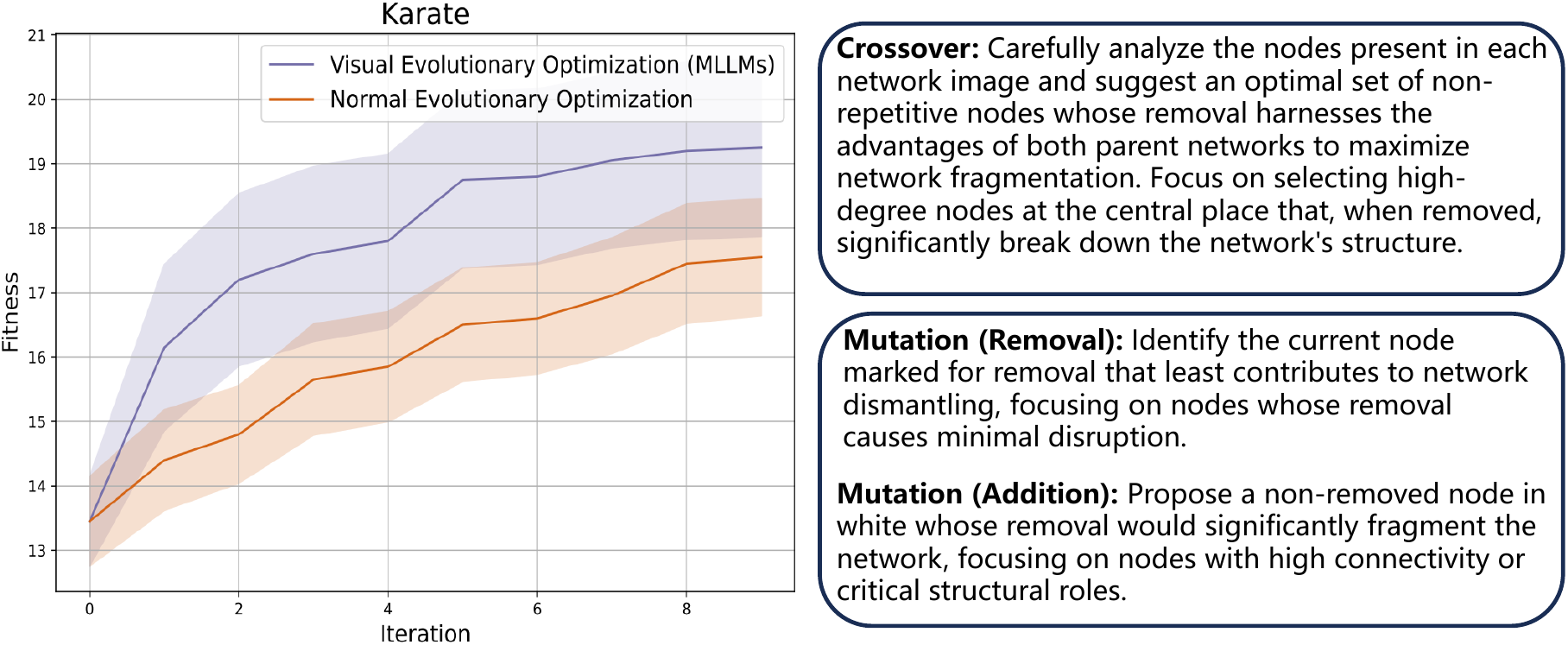}
\caption{{Generalizability test of VEO applied to the network dismantling task on the Karate network.}}
\label{robustness}
\end{figure}

{To investigate the generalizability of VEO, we apply it to the network dismantling problem on the Karate network \cite{artime2024robustness}, as shown in Figure \ref{robustness}. In this task, fitness is defined as the difference between the size of the original network and the size of the largest connected component after node removal. Note that network dismantling is fundamentally different from influence maximization, as the latter focuses on maximizing widespread reach, whereas dismantling aims to disrupt connectivity by targeting structurally critical nodes. The results show that the MLLM-based method consistently outperforms the standard evolutionary strategy across iterations, achieving higher fitness values and faster convergence. The descriptions on the right illustrate how MLLMs are prompted to perform task-specific crossover and mutation operations, confirming the adaptability of the framework to different optimization objectives.} 

\section{{Future work}}\label{sec.future}
{As this work lays the foundation for integrating multimodal large language models (MLLMs) into evolutionary optimization over graph-structured data, several avenues remain open for exploration. We discuss two primary directions that could enhance the effectiveness, scalability, and efficiency of VEO.}

\subsection{{Representation-Level Enhancements}}
\begin{itemize}
\item {\textbf{Adaptive Graph Visualization:} Future efforts may focus on developing visualization techniques that dynamically adapt to graph structure and task context. Highlighting key features such as communities and critical connections could help MLLMs better perceive and reason about graph inputs.}
\item {\textbf{Graph Reduction:} Techniques such as sparsification and abstraction can be further explored to reduce graph complexity while more accurately preserving essential relational patterns, thereby enabling MLLMs to operate more efficiently on large-scale graphs.} \end{itemize}

\subsection{{Optimization-Level Enhancements}}
\begin{itemize}
\item {\textbf{Fine-Tuning MLLMs:} Future work could investigate fine-tuning MLLMs on graph-centric tasks to enhance their adaptability and domain alignment. Fine-tuning lightweight variants may also improve inference speed and resource efficiency.}
\item {\textbf{Advanced Optimization Techniques:} Exploring strategies such as prompt engineering, ensemble approaches, or multi-agent collaboration among MLLMs could further improve solution quality, robustness, and generalization across diverse graph problems.} \end{itemize}

\section{Conclusion}\label{sec.conclusion}
In this paper, we have proposed an original and novel framework named visual evolutionary optimization (VEO) powered by  multimodal large language models (MLLMs). The framework incorporates encoding schemes and MLLM-based evolutionary operators to facilitate intuitive and context-aware adaptations and optimization. This study contributes to the existing body of work on MLLM-driven evolutionary optimization and discusses a viable path for further research in optimizing complex network-based problems. In addition, several key factors to the outcome of VEO have been investigated such as the layout as well as the scale of the simplified network. In future work, we will explore the application of MLLMs to additional graph-structured combinatorial tasks beyond influence maximization by designing relevant prompting strategies and visualization techniques to enhance VEO's applicability and versatility.

\bibliographystyle{IEEEtran}
\bibliography{zhao}

\vfill

\end{document}